\documentclass{article}
\usepackage{spconf,amsmath,graphicx}
\usepackage{epsfig}
\usepackage{makecell}
\usepackage{amsfonts}
\usepackage{hyperref}
\usepackage{xcolor}
\usepackage{subcaption}
\usepackage{multirow}
\usepackage{amssymb}
\usepackage{booktabs}
\usepackage{color, colortbl}

\usepackage{enumitem}

\definecolor{mygray}{gray}{0.9}

\let\OLDthebibliography\thebibliography
\renewcommand\thebibliography[1]{
  \OLDthebibliography{#1}
  \setlength{\parskip}{0pt}
  \setlength{\itemsep}{0pt plus 0.3ex}
}

\title{Multi-level and Multi-modal Action Anticipation}
\name{Seulgi Kim, Ghazal Kaviani, Mohit Prabhushankar, Ghassan AlRegib}

\address {OLIVES at the Center for Signal and Information Processing CSIP,\\ 
School of Electrical and Computer Engineering, Georgia Institute of Technology, Atlanta, GA, USA \\
\{seulgi.kim, gkaviani3, mohit.p, alregib\}@gatech.edu             }

\begin{document}
%%%%%%%%%%%%%%%%%%%
\twocolumn[{%

{ \large
\begin{itemize}[leftmargin=2.5cm, align=parleft, labelsep=2cm, itemsep=4ex,]

\item[\textbf{Citation}]{S. Kim, G. Kaviani, M. Prabhushankar, G. AlRegib, "Multi-level and Multi-modal Action Anticipation," in \textit{2025 IEEE International Conference on Image Processing (ICIP), Anchorage, Alaska, 2025.}}

\item[\textbf{Review}]{Date of Acceptance: May 20th 2025}

\item[\textbf{Codes}]{\url{https://github.com/olivesgatech/mM-ant}}

\item[\textbf{Bib}]  {@inproceedings\{kim2025multilevel,\\
    title=\{Multi-level and Multi-modal Action Anticipation\},\\
    author=\{Kim, Seulgi and Kaviani, Ghazal and Prabhushankar, Mohit and AlRegib, Ghassan\},\\
    booktitle=\{2025 IEEE International Conference on Image Processing (ICIP)\},\\
    year=\{2025\}\}}

%\item[\textbf{Copyright}]{\textcopyright 2022 IEEE. Personal use of this material is permitted. Permission from IEEE must be obtained for all other uses, in any current or future media, including reprinting/republishing this material for advertising or promotional purposes,
%creating new collective works, for resale or redistribution to servers or lists, or reuse of any copyrighted component
%of this work in other works.}

\item[\textbf{Contact}]{
\{seulgi.kim, gkaviani, mohit.p, alregib\}@gatech.edu\\\url{https://alregib.ece.gatech.edu/}\\}
\end{itemize}

}}]

%%%%%%%%%%%%%%%%%%%%
\sloppy
%\ninept
\linespread{0.87}
\maketitle

\begin{figure*}[ht!]
\begin{center}
\includegraphics[width=1.0\linewidth]{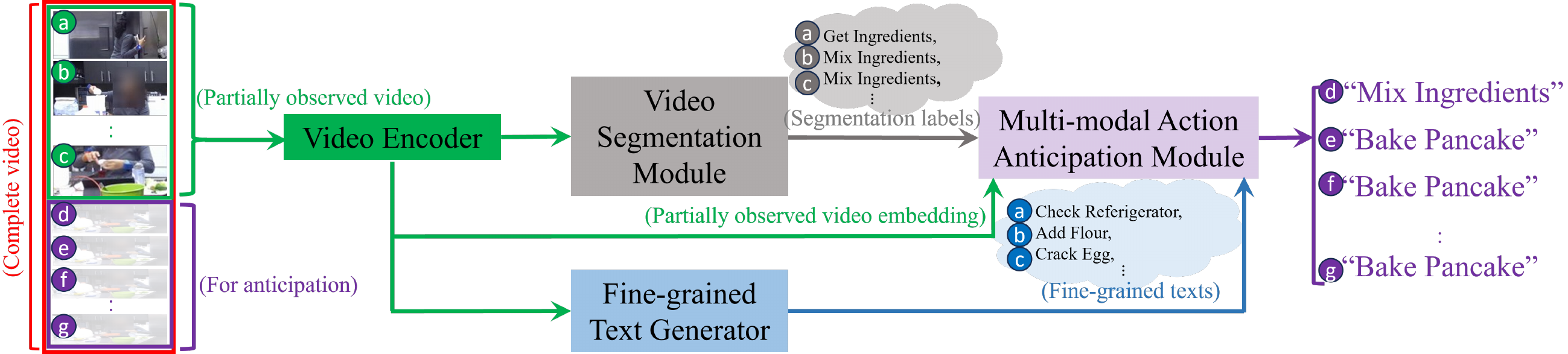}
\end{center}
\vspace{-0.5cm}
   \caption{Flowchart of the \textit{m\&m-Ant} method. Each module and its corresponding input/output are distinguished by color. The green represents the video encoder, which takes a partially observed video as input and outputs embeddings. The gray illustrates the video segmentation module, which generates segmentation labels shown in the gray cloud. The blue represents the fine-grained text generator, which outputs fine-grained text as depicted in the blue cloud. The purple indicates the multi-modal action anticipation module and the anticipation result.}
\label{fig:figure1}
\vspace{-0.5cm}
\end{figure*}

\begin{abstract}
Action anticipation, the task of predicting future actions from partially observed videos, is crucial for advancing intelligent systems. Unlike action recognition, which operates on fully observed videos, action anticipation must handle incomplete information. Hence, it requires temporal reasoning, and inherent uncertainty handling. While recent advances have been made, traditional methods often focus solely on visual modalities, neglecting the potential of integrating multiple sources of information. Drawing inspiration from human behavior, we introduce \textit{Multi-level and Multi-modal Action Anticipation (m\&m-Ant)}, a novel multi-modal action anticipation approach that combines both visual and textual cues, while explicitly modeling hierarchical semantic information for more accurate predictions. To address the challenge of inaccurate coarse action labels, we propose a fine-grained label generator paired with a specialized temporal consistency loss function to optimize performance. Extensive experiments on widely used datasets, including Breakfast, 50 Salads, and DARai, demonstrate the effectiveness of our approach, achieving state-of-the-art results with an average anticipation accuracy improvement of 3.08\% over existing methods. 
This work underscores the potential of multi-modal and hierarchical modeling in advancing action anticipation and establishes a new benchmark for future research in the field. Our code is available at: \href{https://github.com/olivesgatech/mM-ant}{https://github.com/olivesgatech/mM-ant}.

\end{abstract}

\begin{keywords}
Multi-modal action anticipation, Fine-grained text generator, Multi-level (Hierarchical) modeling
\end{keywords}
\section{Introduction}
\label{sec:intro}

Action anticipation is a critical task for advancing intelligent systems, where an algorithm processes a partially observed video to predict a sequence of future actions by utilizing the available visual context \cite{gong2024actfusion,gong2022future,zhang2024object}. In contrast to action recognition \cite{li2024hierarchical,guo2022uncertainty,zhai2023soar}, which focuses on identifying actions from fully observed videos, action anticipation presents unique challenges due to its predictive nature.
First, as illustrated in Figure~\ref{fig:figure1}, the algorithm must operate with incomplete information, relying on a \emph{Partially Observed} frames of the video to predict upcoming actions accurately. This requires the ability to infer event progression from subtle cues within the observed content. For example, in Figure~\ref{fig:figure1}, observing the tasks of ``checking the refrigerator'', ``adding flour'', and ``cracking the eggs'' should be used to infer that the next action is ``mixing ingredients''.
% First, as we see in Figure~\ref{fig:figure1}, the algorithm must operate with incomplete information, relying on a partial segment of the video to predict upcoming actions accurately. This is shown in Figure~\ref{fig:figure1} where the \emph{Partially Observed} frames are used by the algorithm to anticipate the actions in the grayed-out frames in the Daily Activity Recognition (DARai) dataset~\cite{ecnr-hy49-24}. Due to this characteristic, action anticipation requires the ability to infer event progression from subtle cues within the observed content. For example, in Figure~\ref{fig:figure1}, observing the tasks of ``checking the refrigerator'', ``adding flour'', and ``cracking the eggs'' should be used to infer that the next action is ``mixing ingredients''. 
% Second, effective temporal reasoning is essential, as anticipating future actions demands understanding the dynamic relationships between past and upcoming events. 
Second, a single observed scenario can lead to multiple plausible predictions, leading to prediction uncertainty. Instead of ``mixing ingredients'', the human performing the task in Figure~\ref{fig:figure1} may choose to ``switch on the stove''. Together, these challenges make action anticipation challenging, requiring innovative approaches to model temporal dependencies, and interpret contextual signals. 

Recent advancements have proposed various methods to address these challenges \cite{abu2018will,ke2019time,sener2020temporal,abu2021long,gong2022future,gong2024actfusion,zhang2024object,guo2024uncertainty}. However, existing approaches predominantly focus on visual information, often overlooking the potential benefits of integrating other modalities, such as textual data. This reliance on a single modality limits their ability to capture complementary and context-rich information, resulting in suboptimal performance in complex and dynamic environments.

% Human behavior in daily activities demonstrates the importance of multi-modal information in anticipating future actions. For instance, if the recipe used to make the pancake was available to the model in Figure~\ref{fig:figure1}, it would have segmented the subtle details of the partial video frames to anticipate the next action. Hence, multi-modal video and textual inputs allow complementary and context-rich information, resulting in optimal performance in complex and dynamic environments.
Human behavior in daily activities demonstrates the importance of multi-modal information in anticipating future actions. Consider the scenario in Figure \ref{fig:figure1}. One can infer that someone is ``making a pancake'' by observing a sequence of cues such as ``reading a recipe (text),'' ``checking the refrigerator,'' ``adding flour,'' and ``cracking the eggs.'' Similarly, if the recipe used to make the pancake was available to the model in Figure~\ref{fig:figure1}, it would better segment the subtle details of the partial video frames to anticipate the next action. Hence, multi-modal video and textual inputs allow complementary and context-rich information.
Drawing inspiration from human behavior in daily activities, we propose \textit{Multi-level and Multi-modal Action Anticipation (m\&m-Ant)}, to tackle the anticipation task within a multi-modal framework. In the recent works that takes a video segment, segmenting each labels, and predicts future actions, we identify that inaccurate segmentation can mislead the model, leading to suboptimal performance. To address this challenge, we propose a new fine-grained text generator, equipped with a novel temporal consistency loss function, to produce textual outputs that allow action anticipation within a multi-modal framework. The flowchart of our approach is illustrated in Figure \ref{fig:figure1}.

To assess the effectiveness of \textit{m\&m-Ant}, we conduct comprehensive experiments on three widely used datasets: Breakfast \cite{kuehne2014language}, 50 Salads \cite{stein2013combining}, and DARai \cite{kaviani2025hierarchical}. The results demonstrate that our method effectively integrates multi-modal information, surpassing existing approaches and achieving state-of-the-art performance with 3.08\% average improvement in accuracy.
The key contributions of this work are summarized as follows:
\begin{itemize}
    \item \textbf{Proposing a novel multi-modal and hierarchical modeling for action anticipation task}: We introduce an innovative approach, \textit{m\&m-Ant}, that leverages hierarchical and fine-grained semantics by using both visual and textual modalities to predict future actions, addressing the limitations of traditional single-modality approaches.

    \item \textbf{Proposing a specialized loss function for training fine-grained label generator}: We propose a novel temporal consistency loss function that is specifically tailored to address the challenges of temporal coherence and semantic separation in the fine-grained representation space.

    \item \textbf{Empirical validation of the proposed method's effectiveness}: Through comprehensive experiments on benchmark datasets, our method demonstrates superior performance, achieving a state-of-the-art accuracy improvement of 3.08\% over existing approaches.

    % \item \textbf{Introducing a fine-grained label generator and \textcolor{red}{a controller}}: To overcome the challenges posed by inaccurate coarse action labels, a fine-grained label generator with a specially designed fine-grained label controller is proposed, enhancing the model’s ability to make more accurate predictions. \textcolor{red}{Include Loss function as our contribution}
\end{itemize}

The remainder of this paper is organized as follows: Section 2 provides an overview of related work. Section 3 details our proposed method. Finally, Sections 4 and 5 present extensive experimental results and analyze the effectiveness of our approach.

\begin{figure}[t!]
\begin{center}
\includegraphics[width=1.0\linewidth]{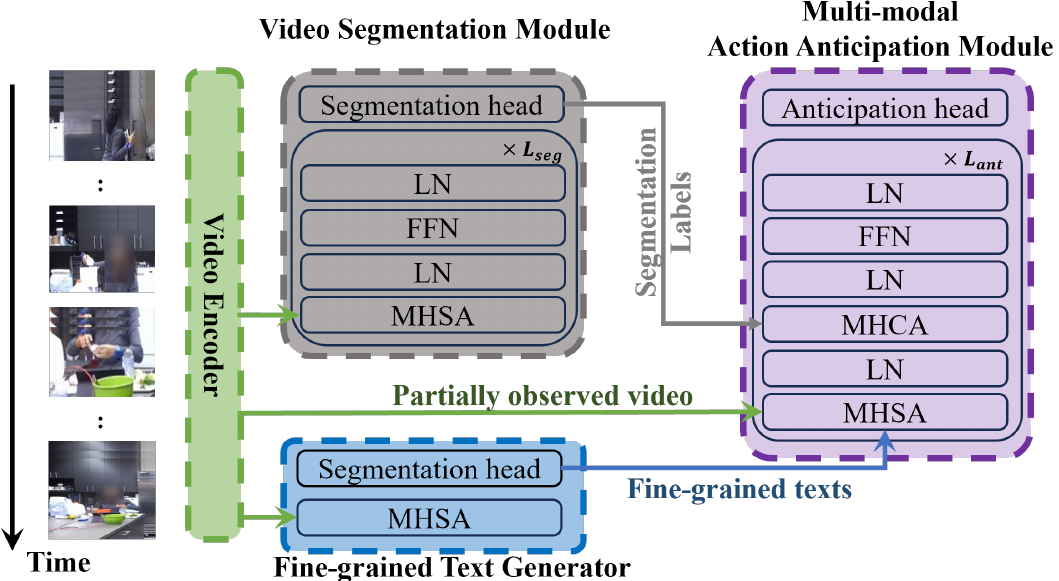}
\end{center}
\vspace{-0.5cm}
   \caption{Detailed architecture of \textit{m\&m-Ant} method. The framework consists of three primary components: the video segmentation module that generates segmentation labels of each frame, the fine-grained text generator for detailed semantic information, and the multi-modal action anticipation module that integrates video and text modalities to predict unseen actions. This design enhances understanding of partially observed videos by leveraging fine-grained information with multi-modal framework. Here, `MHSA' stands for Multi-Head Self-Attention, `LN' for Layer Normalization, `FFN' for Feed-Forward Network, and `MHCA' for Multi-Head Cross-Attention.}
\label{fig:figure2}
\vspace{-0.7cm}
\end{figure}

% \begin{figure}[ht!]
%   \centering
%   \begin{minipage}[t]{0.3\textwidth}
%     \vspace{0pt}
%     \caption{%
%       \textbf{Detailed architecture of \textit{m\&m-Ant} method}. The framework consists of three primary components: the video segmentation module that generates segmentation labels of each frame, the fine-grained text generator for detailed semantic information, and the multi-modal action anticipation module that integrates video and text modalities to predict unseen actions. This design enhances understanding of partially observed videos by leveraging fine-grained information with multi-modal framework. Here, `MHSA' stands for Multi-Head Self-Attention, `LN' for Layer Normalization, `FFN' for Feed-Forward Network, and `MHCA' for Multi-Head Cross-Attention.
%     }
%     \label{fig:figure2}
%   \end{minipage}
%   \hfill
%   \begin{minipage}[t]{0.67\textwidth}
%     \vspace{0pt}
%     \centering
%     \includegraphics[width=\linewidth]{figure/figure2 (cropped) (pdfresizer.com).pdf}
%   \end{minipage}
% \vspace{-0.5cm}
% \end{figure}
\vspace{-0.2cm}
\section{Related Work}
%In this section, we begin by reviewing existing approaches to action understanding, covering key tasks such as action anticipation, early action prediction, and action recognition. Next, we examine prior research on uncertainty in the context of action anticipation.

\subsection{Long-term Action Anticipation}
%Action anticipation is a task that differs from early action prediction and action recognition, despite their shared focus on understanding actions in videos. It involves forecasting future actions before they occur, relying solely on limited observations from a video.

% \subsubsection{Action Anticipation}
%\noindent\textbf{Action anticipation}
Action anticipation emphasizes predicting potential upcoming actions before they are observed. The availability of large-scale video datasets \cite{kaviani2025hierarchical,damen2022rescaling,grauman2022ego4d} has spurred significant progress in addressing the challenge of action anticipation. Action anticipation methods can be broadly categorized into two types: short-term action anticipation and long-term action anticipation. Short-term action anticipation focuses on predicting a single future action that will occur within a few seconds \cite{fernando2021anticipating}. In contrast, long-term action anticipation seeks to forecast an extended sequence of future actions from a long-range video, aiming to predict events far into the future \cite{abu2019uncertainty,abu2018will,abu2021long,sener2020temporal,ke2019time,gong2024actfusion}. In this paper, we tackle the long-term anticipation task. In DARai dataset, this long-term anticipation can be as long as $4$ minutes into the future.

The long-term action anticipation task was first introduced in \cite{abu2018will} where the authors propose a framework that integrates recurrent neural networks (RNNs) with convolutional neural networks (CNNs). Later, \cite{abu2019uncertainty} utilizes gated recurrent units (GRUs) to model uncertainty in future activities autoregressively, enabling the prediction of multiple potential future action sequences during testing. Additionally, \cite{gong2024actfusion} proposes a unified diffusion model capable of addressing both temporal action segmentation and long-term action anticipation within a single architecture and training process. Typically, these methods \cite{abu2018will, abu2019uncertainty, gong2024actfusion} rely on action labels from observed frames, which are extracted using action segmentation models such as \cite{richard2017weakly}. In contrast, recent studies \cite{sener2020temporal, abu2021long} leverage visual features as inputs, utilizing an action segmentation model such as \cite{farha2019ms} to extract these features during training. \cite{sener2020temporal} proposes a multi-scale temporal aggregation model that condenses past visual features into compact vectors and iteratively predicts future actions using a long short-term memory (LSTM) network. Meanwhile, \cite{abu2021long} presents a GRU-based model that enforces cycle consistency between past and future actions to enhance prediction accuracy.
Inspired by the way humans perform action anticipation, we integrate multi-modal information as input in this work, rather than using them separately as in existing methods, to enhance model performance.

\vspace{-0.1cm}
\subsection{Uncertainty for Action Anticipation}
The inherent uncertainty of future events adds complexity to the action anticipation task, as a single observed scenario can yield multiple plausible predictions \cite{furnari2018leveraging,abu2019uncertainty,malinin2018predictive,suris2021learning,guo2024uncertainty}. \cite{abu2019uncertainty} models the probability distribution of future actions, generating multiple samples to account for uncertainty. In \cite{furnari2018leveraging}, the authors incorporate uncertainty into the loss function design. In cases where inferring future actions is particularly challenging, \cite{suris2021learning} proposes a hierarchical model to predict high-level activities. \cite{guo2024uncertainty} models uncertainty in nouns and verbs to identify reliable information, aiding decision-making. Leveraging the existence of fine-grained groupings within the image representation space \cite{kokilepersaud2025hex}, we introduce a fine-grained text generator that first extracts fine-grained visual features and then translates them into semantically rich texts to help disambiguate uncertainty in action anticipation.

\vspace{-0.1cm}
\section{Methodology}
\vspace{-0.1cm}
This section details the components of \textit{m\&m-Ant}. We begin with the Video Encoder, which extracts meaningful features from the input video. The Video Segmentation Module produces initial segmentation labels for each input frame. To address the limitations and uncertainties of these segmentation labels, we introduce the Fine-grained Text Generator, which incorporates detailed semantic information to enhance the representation and accuracy of the labels. Lastly, we describe the Multi-modal Action Anticipation Module, which integrates multi-modal information to enhance the accuracy and reliability of the predicted actions. Each module contributes to addressing the challenges of uncertainty and improving the performance of the action anticipation task. Figure \ref{fig:figure2} provides a detailed illustration of the architecture.

\subsection{Video Encoder}
% We have a dataset consisting of $M$ videos, represented as $\{X_n\}_{n=1}^N$. For input to the encoder, each video $X_n = \{x_i\}_{i=1}^{T}$ are embedded into visual features denoted as $F \in \mathbb{R}^{T \times C} $, where $T$ is the number of frames and $C$ is the feature embedding dimension.
% To reduce computational cost while preserving temporal structure, we sampled frames at regular intervals using a temporal stride $\tau$. This results in a sampled sequence $F_{\tau} \in \mathbb{R}^{T_{\tau} \times C}$, where $T_{\tau} = \left\lfloor \frac{T}{\tau} \right\rfloor$ represents the number of sampled frames.
% The sampled frame features are projected through a linear transformation layer $W \in \mathbb{R}^{C \times D}$, followed by a ReLU activation function:

We have a dataset consisting of $M$ videos, denoted as $\{X_n\}_{n=1}^N$. Each video $X_n = \{x_i\}_{i=1}^{T}$ is embedded into visual features $F \in \mathbb{R}^{T \times C} $, where $T$ is the number of frames and $C$ is the feature embedding dimension. To reduce computational cost while maintaining temporal structure, we sample frames at regular intervals using a temporal stride \( \tau \). This results in a sampled sequence $F_{\tau} \in \mathbb{R}^{T_{\tau} \times C}$, where $T_{\tau} = \left\lfloor \frac{T}{\tau} \right\rfloor$ is the number of sampled frames. 
The sampled frame features are then passed through a linear transformation layer $W \in \mathbb{R}^{C \times D}$, followed by a $\text{ReLU}(\cdot)$ activation function:
\begin{equation}
X_0 = \text{ReLU}(F_{\tau}W),
\end{equation}
where $X_0 \in \mathbb{R}^{T_{\tau} \times D}$ represents the input tokens and $D$ is the new feature dimension.

%4.5475, 3.45, 1.045
%8.0775, 11.08, -3.3575
%1.5825, 0.2475, 1.0125

\subsection{Fine-grained Text Generator}
% The standard classification loss treat the two frames with the same action label as identical. However, in practice, the same frames occurring at different timestamps can be part of entirely separate sub-action flows. At the same time, frames captured in close temporal proximity exhibit highly similar appearances.
% To address these nuances, we introduce a novel \textit{Temporal Consistency Loss} that considers both class and time. Specifically, we first group frames belonging to the same action class and lyung within a continuous temporal interval into a single cluster. We then apply an \textit{intra-cluster consistency loss} to ensure that frames within the same cluster remain closely aligned. Given the features $x_i$, we minimize the mean squared error between each feature $x_i$ in cluster $k$ and their corresponding cluster mean $\mu_k$:

Standard classification loss treats frames with the same action label as identical. However, in practice, frames with the same label occurring at different timestamps may belong to distinct sub-action flows, while frames captured in close temporal proximity tend to have highly similar appearances.

\noindent\textbf{Temporal consistency loss.} To tackle these nuances, we propose a novel temporal consistency loss that takes both class and time into account. \\
We first group frames from the same action class that lie within a continuous temporal interval into a single cluster. Then, we apply an \textit{intra-cluster consistency loss} to ensure frames within the same cluster remain closely aligned. Given the features $x_i$, we minimize the mean squared error between each feature $x_i$ in cluster $k$ and the corresponding cluster mean $\mu_k$:
$\mathcal{L}_{\text{intra}} = \sum_{k} \sum_{x_{i} \in X(k)} \|x_{i} - \mu_k\|^2,$
where $\mu_k = \frac{1}{|X(k)|} \sum_{x_{i} \in X(k)} x_{i}$ indicates the centroid of cluster $k$ and $X(k)$ denotes the set of features in cluster $k$.
Additionally, we encourage frames that are temporally distant to be assigned to different clusters, even if they belong to the same class label. To achieve this, we introduce an \textit{inter-cluster separation loss}:
$\mathcal{L}_{\text{inter}} = \sum_{k, k' \neq k} \frac{1}{\| \mu_k - \mu_{k'} \|},$
where $\| \mu_k - \mu_{k'} \|$ is the Euclidean distance between the centroids of clusters $k$ and $k'$. 
This term penalizes clusters that are too close in the embedding space, ensuring that repeated instances of the same action occurring at widely separated time intervals are placed in different clusters.
% where \(\| \mu_k - \mu_{k'} \|\) is the Euclidean distance between the centroids of cluster \(k\) and cluster \(k'\). This term penalizes clusters that are too close to each other in the embedding space. By doing this, the repeated instances of the same action at widely separated time intervals are placed in different clusters.
Finally, our proposed temporal consistency loss is formulated as the sum of the \textit{intra-cluster consistency loss} and \textit{inter-cluster separation loss}:
\begin{equation}
\mathcal{L}_{\text{tcl}} = \lambda_1 \mathcal{L}_{\text{intra}} + \lambda_2 \mathcal{L}_{\text{inter}},
\end{equation}
where ``tcl'' represents the temporal consistency loss, and $\lambda_1$ and $\lambda_2$ are weighting factors that control the balance between intra-cluster cohesion and inter-cluster separation. \\
The proposed temporal consistency loss enables the fine-grained text generator to more effectively capture subtle differences in sub-actions over time.

\noindent\textbf{Fine-grained text generator training.} To train the fine-grained text generator, this clustering loss is combined with the cross-entropy (ce) loss for the fine-grained labels:
\begin{equation}
\mathcal{L}_{\text{total}} = \mathcal{L}_{\text{ce}} + \mathcal{L}_{\text{tcl}}.
\end{equation}

\subsection{Video Segmentation Module}
The video segmentation module takes video features as input and produces action labels for each frame. It consists of three main components: a stack of $N$ layers of multi-head self-attention and feed-forward networks, followed by a final fully connected layer. The structure is detailed as follows:
% Coarse label generator takes video features as an input and generates coarse labels that corresponds to each input frames. The coarse label generator consists of three main components: a stack of $N$ layers of multi-head self-attention and feed-forward networks, and a final fully connected layer at the end. The structure is detailed as follows: \\
% \vspace{0.01cm}

\noindent \textbf{Multi-head self-attention.} 
In the video segmentation module, each layer starts with multi-head self-attention, which captures contextual dependencies within the input sequence. For an input sequence representation $\mathbf{H}^{(l)} \in \mathbb{R}^{T \times d}$ at the $l^{th}$ layer, the multi-head self-attention $\text{MHSA}(\cdot)$ is defined as:
$\text{MHSA}(\mathbf{H}^{(l)}) = \text{Concat}(\mathbf{A}_1, \mathbf{A}_2, \dots, \mathbf{A}_h) \mathbf{W}_o,$
where $\text{Concat}(\cdot)$ denotes a concatenation function, $h$ is the number of attention heads, and $\mathbf{W}_o \in \mathbb{R}^{(h \cdot d_h) \times d}$ is the output projection matrix. 

The computation for each attention head is formulated as follows:
\begin{equation}
\mathbf{A}_i = \text{Softmax}\left(\frac{\mathbf{H}^{(l)} \mathbf{W}_i^Q (\mathbf{H}^{(l)} \mathbf{W}_i^K)^\top}{\sqrt{d_h}}\right)(\mathbf{H}^{(l)} \mathbf{W}_i^V),
\end{equation}
where $\text{Softmax}(\cdot)$ refers to the softmax function, and $\mathbf{W}_i^Q, \mathbf{W}_i^K, \mathbf{W}_i^V \in \mathbb{R}^{d \times d_h}$ are the query, key, and value projection matrices, respectively. The same input $\mathbf{H}^{(l)}$ is used for queries, keys, and values to implement the self-attention mechanism. Finally, layer normalization (LN) is applied to the output to stabilize training.
% where $\mathbf{W}_i^Q, \mathbf{W}_i^K, \mathbf{W}_i^V \in \mathbb{R}^{d \times d_h}$ are query, key, and value projections. Here, the same input $\mathbf{H}^{(l)}$ is used for queries, keys, and values, to make a self-attention mechanism. In the end, to stabilize training, layer normalization is applied to the output.

% \vspace{0.01cm}

\noindent \textbf{Feed-forward network.}
The output of $\text{MHSA}(\cdot)$ is passed to a feed-forward network $\text{FFN}(\cdot)$, defined as:
$\text{FFN}(\mathbf{X}) = \sigma(\mathbf{X} \mathbf{W}_1 + \mathbf{b}_1) \mathbf{W}_2 + \mathbf{b}_2,$
where $\mathbf{W}_1 \in \mathbb{R}^{d \times d_f}$, $\mathbf{W}_2 \in \mathbb{R}^{d_f \times d}$, and $\sigma(\cdot)$ is a non-linear activation. 
To enhance stability during training, the residual connection and layer normalization $\text{LN}(\cdot)$ are applied:
\begin{equation}
\mathbf{H}^{(l+1)} = \text{LN}(\text{FFN}(\text{LN}(\text{MHSA}(\mathbf{H}^{(l)} + \mathbf{P})) + \mathbf{H}^{(l)})) + \mathbf{H}^{(l)},
\end{equation}
where $\mathbf{P} \in \mathbb{R}^{T \times d}$ denotes the positional encodings, which capture the sequential order of the input frames.

% \vspace{0.05cm}

\noindent \textbf{Segmentation head.}  The final layer output $\mathbf{H}^{(N)}$ is used to predict the video labels $\hat{\mathbf{Y}}$ via a fully connected layer $\text{FC}(\cdot)$ followed by a softmax:
\begin{equation}
\hat{\mathbf{Y}} = \text{Softmax}(\text{FC}(\mathbf{H}^{(N)})).
\end{equation}

\subsection{Multi-modal Action Anticipation Module}
% Our goal is to reason the future actions by modeling temporal dependencies and semantic relationships using hierarchical information. Multi-modal action anticipation module takes inputs by concatenating video feature and corresponding fine-grained labels. In other words:
Our goal is to anticipate future actions by capturing temporal dependencies and semantic relationships through hierarchical modeling. The multi-modal action anticipation module achieves this by integrating video features with their corresponding fine-grained texts. Specifically:
$\mathbf{H}^{\text{input}} = \text{Concat}(\mathbf{H}^{\text{video}}, \mathbf{H}^{\text{fine-grained}}),$
where \(\mathbf{H}^{\text{video}} \in \mathbb{R}^{T \times d}\), \(\mathbf{H}^{\text{fine-grained}} \in \mathbb{R}^{T \times d}\), and \(T\) is the sequence length.
% By feeding $\mathbf{H}^{\text{input}}$ into the $\text{MHSA}(\cdot)$, the model discovers direct correspondences between fine-grained labels' detailed semantics and the visual cues embedded in the image features. Then, the multi-head cross-attention module $\text{MHCS}(\cdot)$ aligns coarse level semantics with the output of MHSA:
By inputting $\mathbf{H}^{\text{input}}$ into the multi-head self-attention, the model learns to capture direct correspondences between the detailed semantics of fine-grained labels and the visual cues embedded in the video features. Subsequently, the multi-head cross-attention module $\text{MHCS}(\cdot)$ explicitly incorporates high-level semantic information from the output of Video Segmentation Module. Video labels $\mathbf{H}^{\text{vidSeg}} \in \mathbb{R}^{T \times d}$ act as keys and values in the attention mechanism, aligning them with the output of multi-head self-attention ($\mathbf{H}^{\text{MHSA}}$):
\begin{equation}
\text{MHCS}(\mathbf{Q}, \mathbf{K}, \mathbf{V}) = \text{Softmax}\left(\frac{\mathbf{Q} \mathbf{K}^\top}{\sqrt{d_h}}\right) \mathbf{V},
\end{equation}
where the output of ($\mathbf{H}^{\text{MHSA}}$) is used as the queries $\mathbf{Q}$, while the video labels $\mathbf{H}^{\text{vidSeg}}$ serve as both keys $\mathbf{K}$ and values $\mathbf{V}$.

\section{Experiments and Analysis}
In this section, we provide an overview of the datasets and experimental setups employed in our experiments, followed by detailed analyses of the proposed method.

\vspace{-0.1cm}
\subsection{Datasets and Implementation Details}

\textbf{Datasets.}
We aim to evaluate our model's ability to handle diverse environments, sparse temporal patterns, and dense action sequences. To do this, we make use of three action anticipation datasets: Breakfast \cite{kuehne2014language}, 50 Salads \cite{stein2013combining}, and DARai \cite{kaviani2025hierarchical}. The Breakfast dataset is one of the largest and most diverse benchmarks for action anticipation, comprising 1,712 videos of 52 individuals, 18 distinct kitchen environments, with 10 coarse-level labels and 48 fine-grained labels. The 50 Salads dataset focuses on fine-grained action segmentation within longer and denser activity sequences. For example, unlike Breakfast which has around 6 actions per video, the videos in 50 Salads are more temporally dense and more transitions, averaging around 20 actions per video. It consists of 50 videos, 25 participants, 17 detailed action labels and 3 broader activity categories. The DARai dataset offers a highly realistic scenarios that closely resemble real-world human behavior. Unlike Breakfast and 50 Salads, the videos in DARai are untrimmed, showcasing raw, continuous human activity in real-world contexts without artificial segmentation. It comprises 150 action classes, 50 participants, two distinct exocentric views, and three levels of hierarchical label.

\begin{table}[ht]
\centering
\begin{footnotesize}
\renewcommand{\arraystretch}{1.2} % Adjust row spacing slightly
\scalebox{0.68}{
\begin{tabular}{cccccccccc}
\hline
Dataset      & Methods   & \multicolumn{4}{c}{$\alpha=0.2$}                & \multicolumn{4}{c}{$\alpha=0.3$}                 \\ \cline{3-10} 
             &           & $\beta(0.1)$ & $\beta(0.2)$ & $\beta(0.3)$ & $\beta(0.5)$ & $\beta(0.1)$ & $\beta(0.2)$ & $\beta(0.3)$ & $\beta(0.5)$ \\ \hline
Breakfast    & RNN \cite{abu2018will}   & 18.11        & 17.20        & 15.94        & 15.81        & 21.64        & 20.02        & 19.73        & 19.21        \\
             & CNN \cite{abu2018will}   & 17.90        & 16.35        & 15.37        & 14.54        & 22.44        & 20.12        & 19.69        & 18.76        \\
             & FUTR \cite{gong2022future}  & 47.06        & 47.08        & 47.11        & 47.11        & 65.55        & 65.55        & 65.54        & 65.51        \\
             & GTD \cite{zatsarynna2024gated}  & 49.86        & 49.75        & 49.65        & 49.58        & 65.00        & 64.39        & 63.95        & 63.79        \\
             & \textbf{m\&m-Ant}  & \textbf{50.55} & \textbf{50.53} & \textbf{50.52} & \textbf{50.56} & \textbf{66.49} & \textbf{66.53} & \textbf{66.63} & \textbf{66.68} \\ \hline
50 Salads    & RNN \cite{abu2018will}   & 30.06        & 25.43        & 18.74        & 13.49        & 30.77        & 17.19        & 14.79        & 09.77        \\
             & CNN \cite{abu2018will}   & 21.24        & 19.03        & 15.98        & 09.87        & 29.14        & 20.14        & 17.46        & 10.86        \\
             & FUTR \cite{gong2022future}  & 54.83        & 54.33        & 52.45        & \textbf{51.03}        & 100.00       & \textbf{100.00}        & 98.62        & \textbf{53.00}        \\
             & \textbf{m\&m-Ant}  & \textbf{69.61} & \textbf{69.27} & \textbf{69.73} & 48.35 & \textbf{100.00} & 99.88 & \textbf{98.70} & 39.61 \\ \hline
\end{tabular}}
\end{footnotesize}
\vspace{-0.2cm}
\caption{State-of-the-art performance comparison on Breakfast and 50Salads datasets with varying $\beta$ values for $\alpha=0.2$ and $\alpha=0.3$.}
\vspace{-0.2cm}
\label{tab:breakfast_50salads}
\end{table}

\begin{table}[ht]
\centering
\renewcommand{\arraystretch}{1.3} % Adjust row spacing to make it less tight
\scalebox{0.68}{
\begin{tabular}{cccccc}
\hline
$\alpha$ & Methods   & $\beta(0.1)$ & $\beta(0.2)$ & $\beta(0.3)$ & $\beta(0.5)$ \\ \hline
0.1      & FUTR \cite{gong2022future}  & 24.26        & 24.25        & 24.74        & 23.46        \\
& AFFT \cite{zhong2023anticipative}  & 20.25        & 23.13        & 23.63        & 23.42        \\
         & \textbf{m\&m-Ant}  & \textbf{26.53} & \textbf{25.82} & \textbf{25.98} & \textbf{24.71} \\ \hline
0.2      & FUTR \cite{gong2022future}  & 25.05        & 25.11        & \textbf{24.48}        & \textbf{23.18}        \\
& AFFT \cite{zhong2023anticipative}  & 23.14        & 24.78        & 23.62        & 21.02        \\
         & \textbf{m\&m-Ant}  & \textbf{25.75} & \textbf{25.70} & 24.24 & 23.12 \\ \hline
0.3      & FUTR \cite{gong2022future}  & 40.71        & 33.57        & 33.42        & 30.79        \\
& AFFT \cite{zhong2023anticipative}  & 33.82        & 29.25        & 28.33        & 25.45        \\
         & \textbf{m\&m-Ant}  & \textbf{42.00} & \textbf{34.71} & \textbf{34.49} & \textbf{31.34} \\ \hline
\end{tabular}}
\vspace{-0.2cm}
\caption{Performance comparison on DARai dataset with varying $\alpha$ and $\beta$ values.}
\vspace{-0.5cm}
\label{tab:darai_results}
\end{table}

\noindent\textbf{Experimental setups.} 
We use pre-extracted ResNet features as input visual features for the Breakfast, 50 Salads, and DARai. To align with the temporal resolution of each dataset, the temporal stride $\tau$ is set to 3 for Breakfast, 6 for 50 Salads, and 15 for DARai. Our model architecture consists of two layers of video segmentation module ($L_{seg}=2$) and one layer of multi-modal action anticipation module ($L_{ant}=1$), with the number of queries fixed at 8. Here, $t$ number of fine-grained labels are aggregated using a pooling layer, where $t$ is the number of input frames. Based on the density of each dataset, the hidden dimension size $D$ is set to 128 for Breakfast and DARai, and 512 for 50 Salads. During training, the observation rate $\alpha$ is set to $\alpha \in \{0.2, 0.3, 0.5\}$, while the prediction rate $\beta$ is fixed at 0.5. The model is trained for 60 epochs using the AdamW optimizer \cite{loshchilov2017decoupled} with a learning rate of $1e-3$ and a batch size of 16. A cosine annealing warm-up scheduler \cite{loshchilov2016sgdr} is applied, with warm-up stages spanning 10 epochs. For evaluation, we set the observation rate $\alpha \in \{0.1, 0.2, 0.3\}$ and prediction rate $\beta \in \{0.1, 0.2, 0.3, 0.5\}$. We measure mean over classes (MoC) accuracy, following the long-term action anticipation framework protocol \cite{abu2018will,abu2021long,sener2020temporal,ke2019time}. To ensure consistency, we report average performance across 3 number of iteration, each with fixed seeds $1, 10, 13452$.

\vspace{-0.2cm}
\subsection{Quantitative Analysis}
\noindent\textbf{State-of-the-art comparison.}
As we observe in Table \ref{tab:breakfast_50salads} and \ref{tab:darai_results}, \textit{m\&m-Ant} consistently outperforms the state-of-the-art across all datasets and evaluation settings, achieving an average accuracy gain of 3.08\%. In Table \ref{tab:breakfast_50salads}, we compare the performance of multiple models and observe that \textit{m\&m-Ant} achieves the best performance among all models, with FUTR serving as the state-of-the-art baseline. This gain is particularly notable in scenarios with lower observation rates ($\alpha = 0.1, 0.2$), where fine-grained textual information allows the model to leverage subtle cues even when visual input is limited. The performance boost is especially pronounced on the 50 Salads dataset, which contains dense action sequences and frequent transitions—here, fine-grained textual information enhances the model’s ability to capture nuanced temporal variations, leading to greater improvements than in other datasets. Table \ref{tab:darai_results} specifically compares FUTR and \textit{m\&m-Ant} on the DARai dataset, which includes untrimmed videos and complex real-world scenarios. Despite the increased complexity of DARai’s second-level hierarchical structure, \textit{m\&m-Ant} consistently achieves higher performance compared to FUTR. Although the improvements on DARai are more moderate due to its complexity, \textit{m\&m-Ant} achieves a maximum accuracy gain of 2.27\%, demonstrating its robustness in handling challenging real-world data.
% As we observe in Table \ref{tab:performance}, \textit{m\&m-Ant} consistently outperforms the state-of-the-art across all datasets and evaluation settings, achieving average accuracy gain of 3.08\%. This gain is particularly notable in scenarios with lower observation rates ($\alpha = 0.1, 0.2$), where fine-grained textual information allows the model to leverage subtle cues even when visual input is limited. The performance boost is especially pronounced on the 50 Salads dataset, which contains dense action sequences and frequent transitions—here, fine-grained textual information enhances the model’s ability to capture nuanced temporal variations, leading to greater improvements than in other datasets.
% For the DARai dataset, which contains untrimmed videos and complex real-world scenarios, our method also consistently outperforms the state-of-the-art method, i.e., FUTR. However, due to the increased complexity of DARai’s second-level hierarchical structure, the improvements are more moderate, with a maximum accuracy gain of 2.27\%.

\noindent\textbf{Ablation study.}
We perform ablation studies to assess the impact of multi-modal (Table \ref{tab:multi_modal_comparison}) and multi-level (hierarchical) modeling (Table \ref{tab:multi_modal_ hierarchica_comparison}). First, we compare a uni-modal model—trained only on visual features with a multi-modal model that uses both visual and textual inputs (i.e., video labels and fine-grained texts). Our results clearly show that the multi-modal approach consistently outperforms the uni-modal counterpart across all observation rates. Second, we evaluate the effect of multi-level (hierarchical) modeling by comparing a model that uses only video labels with one that incorporates both video labels and fine-grained texts. Our results reveal that leveraging fine-grained texts leads to substantial performance gain, particularly at lower observation rates.

\begin{table}[t!]
\centering
\scalebox{0.68}{
\begin{tabular}{llcccc}
\toprule
\textbf{Dataset} & \textbf{Methods} & \multicolumn{4}{c}{\textbf{Observation Rate (Input)}} \\
\cmidrule(lr){3-6}
 &  & 0.1 & 0.2 & 0.3 & 0.4 \\
\midrule
\multirow{2}{*}{Breakfast} & Uni-modal & 26.86 & 47.21 & 62.43 & 66.64 \\
 & Multi-modal & \textbf{30.05} & \textbf{50.54} & \textbf{66.59} & \textbf{67.20} \\
\midrule
\multirow{2}{*}{50 Salads} & Uni-modal & 29.76 & 53.76 & \textbf{90.92} & 73.92 \\
 & Multi-modal & \textbf{39.33} & \textbf{64.24} & 84.55 & \textbf{79.82} \\
\midrule
\multirow{2}{*}{DARai} & Uni-modal & \textbf{25.87} & 22.02 & 28.92 & 27.08\\
 & Multi-modal & 25.76 & \textbf{24.70} & \textbf{35.64} & \textbf{38.36}\\
\bottomrule
\end{tabular}}
\caption{An ablation study comparing uni-modal and multi-modal settings across different datasets and observation rates.}
\vspace{-0.1cm}
\label{tab:multi_modal_comparison}
\end{table}

\begin{table}[t!]
\centering
\scalebox{0.68}{
\begin{tabular}{llcccc}
\toprule
\textbf{Dataset} & \textbf{Methods} & \multicolumn{4}{c}{\textbf{Observation Rate (Input)}} \\
\cmidrule(lr){3-6}
 &  & 0.1 & 0.2 & 0.3 & 0.4 \\
\midrule
\multirow{2}{*}{Breakfast} & w/o multi-level modeling & 25.51 & 47.10 & 65.54 & \textbf{68.24}\\
 & w/ multi-level modeling & \textbf{30.05} & \textbf{50.54} & \textbf{66.59} & 67.20\\
\midrule
\multirow{2}{*}{50 Salads} & w/o multi-level modeling & 31.25 & 53.16 & 87.91 & 68.80\\
 & w/ multi-level modeling & \textbf{39.33} & \textbf{64.24} & \textbf{84.55} & \textbf{79.82} \\
\midrule
\multirow{2}{*}{DARai} & w/o multi-level modeling & 24.18 & 24.45 & 34.63 & 34.16 \\
 & w/ multi-level modeling & \textbf{25.76} & \textbf{24.70} & \textbf{35.64} & \textbf{38.36}\\
\bottomrule
\end{tabular}}
\vspace{-0.3cm}
\caption{An ablation study comparing models with (w/) and without (w/o) multi-level (hierarchical) modeling (fine-grained text generator) across different datasets and observation rates.}
\vspace{-0.6cm}
\label{tab:multi_modal_ hierarchica_comparison}
\end{table}

% \begin{table}
% \begin{subtable}[t]{1.0\linewidth}
% \small
% \centering
% \scalebox{0.7}{
% \begin{tabular}{c|ccccccc}
% \toprule
% \multicolumn{2}{c|}{\textbf{Model}}    & \textbf{Top-1 Acc} & \textbf{Top-5 Acc} & \textbf{Params (M)} & \textbf{FLOPs (G)} & \textbf{Image size} \\ 
% \midrule
% \multicolumn{2}{c|}{AViT-S \cite{yin2022vit}}  & 78.8 & 93.9	& 22 & 3.6  & 224x224 \\ 
% \midrule
% \multicolumn{2}{c|}{EViT-S \cite{liang2022not}}    & 79.7	& 94.9	& 22.1	& 4	& 224x224 \\ 
% \midrule
% \rowcolor{mygray} \multicolumn{2}{c|}{DeiT-S \cite{touvron2021training}}      & 79.8	& 94.9	& 22  & 4.6	& 224x224  \\ 
% \midrule
% \multicolumn{2}{c|}{\textbf{Ours}}       & \textbf{80.2}	& \textbf{95.8}	& \textbf{22}	& \textbf{3.6}	& \textbf{224x224} \\ 
% \bottomrule
% \end{tabular}}
% \caption{State-of-the-art comparisons based on the DARai dataset.}
% \label{table:2-2}
% \end{subtable}
% \end{table}

\vspace{-0.2cm}
\section{Conclusion}
\vspace{-0.2cm}
In conclusion, we introduce \textit{m\&m-Ant}, a novel multi-modal action anticipation approach that effectively integrates both visual and textual information to predict future actions from partially observed videos. By incorporating a fine-grained text generator with a novel loss function, our method addresses the challenges of inaccurate video labels, leading to improved prediction accuracy. Comprehensive experiments on standard datasets demonstrate the superiority of our approach, achieving state-of-the-art performance with a significant accuracy boost. This study underscores the potential of multi-level and multi-modal modeling in enhancing action anticipation and sets a new benchmark for future research.

\bibliographystyle{IEEEbib}
\bibliography{strings,refs}

\begin{thebibliography}{10}

\bibitem{gong2024actfusion}
Dayoung Gong, Suha Kwak, and Minsu Cho,
\newblock ``Actfusion: a unified diffusion model for action segmentation and anticipation,''
\newblock {\em arXiv:2412.04353}, 2024.

\bibitem{gong2022future}
Dayoung Gong, Joonseok Lee, Manjin Kim, Seong~Jong Ha, and Minsu Cho,
\newblock ``Future transformer for long-term action anticipation,''
\newblock in {\em CVPR}, 2022, pp. 3052--3061.

\bibitem{zhang2024object}
Ce~Zhang, Changcheng Fu, Shijie Wang, Nakul Agarwal, Kwonjoon Lee, Chiho Choi, and Chen Sun,
\newblock ``Object-centric video representation for long-term action anticipation,''
\newblock in {\em WACV}, 2024, pp. 6751--6761.

\bibitem{li2024hierarchical}
Changzhen Li, Jie Zhang, Shuzhe Wu, Xin Jin, and Shiguang Shan,
\newblock ``Hierarchical compositional representations for few-shot action recognition,''
\newblock {\em CVIU}, 2024.

\bibitem{guo2022uncertainty}
Hongji Guo, Zhou Ren, Yi~Wu, Gang Hua, and Qiang Ji,
\newblock ``Uncertainty-based spatial-temporal attention for online action detection,''
\newblock in {\em ECCV}. Springer, 2022, pp. 69--86.

\bibitem{zhai2023soar}
Yuanhao Zhai, Ziyi Liu, Zhenyu Wu, Yi~Wu, Chunluan Zhou, David Doermann, Junsong Yuan, and Gang Hua,
\newblock ``Soar: Scene-debiasing open-set action recognition,''
\newblock in {\em ICCV}, 2023, pp. 10244--10254.

\bibitem{abu2018will}
Yazan Abu~Farha, Alexander Richard, and Juergen Gall,
\newblock ``When will you do what?-anticipating temporal occurrences of activities,''
\newblock in {\em CVPR}, 2018, pp. 5343--5352.

\bibitem{ke2019time}
Qiuhong Ke, Mario Fritz, and Bernt Schiele,
\newblock ``Time-conditioned action anticipation in one shot,''
\newblock in {\em CVPR}, 2019, pp. 9925--9934.

\bibitem{sener2020temporal}
Fadime Sener, Dipika Singhania, and Angela Yao,
\newblock ``Temporal aggregate representations for long-range video understanding,''
\newblock in {\em ECCV}. Springer, 2020.

\bibitem{abu2021long}
Yazan Abu~Farha, Qiuhong Ke, Bernt Schiele, and Juergen Gall,
\newblock ``Long-term anticipation of activities with cycle consistency,''
\newblock in {\em PR}. Springer, 2021, pp. 159--173.

\bibitem{guo2024uncertainty}
Hongji Guo, Nakul Agarwal, Shao-Yuan Lo, Kwonjoon Lee, and Qiang Ji,
\newblock ``Uncertainty-aware action decoupling transformer for action anticipation,''
\newblock in {\em CVPR}, 2024, pp. 18644--18654.

\bibitem{kuehne2014language}
Hilde Kuehne, Ali Arslan, and Thomas Serre,
\newblock ``The language of actions: Recovering the syntax and semantics of goal-directed human activities,''
\newblock in {\em CVPR}, 2014.

\bibitem{stein2013combining}
Sebastian Stein and Stephen~J McKenna,
\newblock ``Combining embedded accelerometers with computer vision for recognizing food preparation activities,''
\newblock in {\em Proceedings of the 2013 ACM international joint conference on Pervasive and ubiquitous computing}, 2013, pp. 729--738.

\bibitem{kaviani2025hierarchical}
Ghazal Kaviani, Yavuz Yarici, Seulgi Kim, Mohit Prabhushankar, Ghassan AlRegib, Mashhour Solh, and Ameya Patil,
\newblock ``Hierarchical and multimodal data for daily activity understanding,''
\newblock {\em arXiv preprint arXiv:2504.17696}, 2025.

\bibitem{damen2022rescaling}
Dima Damen, Hazel Doughty, Giovanni~Maria Farinella, Antonino Furnari, Evangelos Kazakos, Jian Ma, Davide Moltisanti, Jonathan Munro, Toby Perrett, Will Price, et~al.,
\newblock ``Rescaling egocentric vision: Collection, pipeline and challenges for epic-kitchens-100,''
\newblock {\em IJCV}, pp. 1--23, 2022.

\bibitem{grauman2022ego4d}
Kristen Grauman, Andrew Westbury, Eugene Byrne, Zachary Chavis, Antonino Furnari, Rohit Girdhar, Jackson Hamburger, Hao Jiang, Miao Liu, Xingyu Liu, et~al.,
\newblock ``Ego4d: Around the world in 3,000 hours of egocentric video,''
\newblock in {\em CVPR}, 2022, pp. 18995--19012.

\bibitem{fernando2021anticipating}
Basura Fernando and Samitha Herath,
\newblock ``Anticipating human actions by correlating past with the future with jaccard similarity measures,''
\newblock in {\em CVPR}, 2021, pp. 13224--13233.

\bibitem{abu2019uncertainty}
Yazan Abu~Farha and Juergen Gall,
\newblock ``Uncertainty-aware anticipation of activities,''
\newblock in {\em ICCVW}, 2019, pp. 0--0.

\bibitem{richard2017weakly}
Alexander Richard, Hilde Kuehne, and Juergen Gall,
\newblock ``Weakly supervised action learning with rnn based fine-to-coarse modeling,''
\newblock in {\em CVPR}, 2017, pp. 754--763.

\bibitem{farha2019ms}
Yazan~Abu Farha and Jurgen Gall,
\newblock ``Ms-tcn: Multi-stage temporal convolutional network for action segmentation,''
\newblock in {\em CVPR}, 2019, pp. 3575--3584.

\bibitem{furnari2018leveraging}
Antonino Furnari, Sebastiano Battiato, and Giovanni Maria~Farinella,
\newblock ``Leveraging uncertainty to rethink loss functions and evaluation measures for egocentric action anticipation,''
\newblock in {\em ECCVW}, 2018, pp. 0--0.

\bibitem{malinin2018predictive}
Andrey Malinin and Mark Gales,
\newblock ``Predictive uncertainty estimation via prior networks,''
\newblock {\em Advances in neural information processing systems}, vol. 31, 2018.

\bibitem{suris2021learning}
D{\'\i}dac Sur{\'\i}s, Ruoshi Liu, and Carl Vondrick,
\newblock ``Learning the predictability of the future,''
\newblock in {\em CVPR}, 2021, pp. 12607--12617.

\bibitem{kokilepersaud2025hex}
Kiran Kokilepersaud, Seulgi Kim, Mohit Prabhushankar, and Ghassan AlRegib,
\newblock ``Hex: Hierarchical emergence exploitation in self-supervised algorithms,''
\newblock in {\em 2025 IEEE/CVF Winter Conference on Applications of Computer Vision (WACV)}. IEEE, 2025, pp. 1111--1121.

\bibitem{zatsarynna2024gated}
Olga Zatsarynna, Emad Bahrami, Yazan~Abu Farha, Gianpiero Francesca, and Juergen Gall,
\newblock ``Gated temporal diffusion for stochastic long-term dense anticipation,''
\newblock in {\em European Conference on Computer Vision}. Springer, 2024, pp. 454--472.

\bibitem{zhong2023anticipative}
Zeyun Zhong, David Schneider, Michael Voit, Rainer Stiefelhagen, and J{\"u}rgen Beyerer,
\newblock ``Anticipative feature fusion transformer for multi-modal action anticipation,''
\newblock in {\em Proceedings of the IEEE/CVF Winter Conference on Applications of Computer Vision}, 2023, pp. 6068--6077.

\bibitem{loshchilov2017decoupled}
Ilya Loshchilov,
\newblock ``Decoupled weight decay regularization,''
\newblock {\em arXiv preprint arXiv:1711.05101}, 2017.

\bibitem{loshchilov2016sgdr}
Ilya Loshchilov and Frank Hutter,
\newblock ``Sgdr: Stochastic gradient descent with warm restarts,''
\newblock {\em arXiv preprint arXiv:1608.03983}, 2016.

\end{thebibliography}

\end{document}